\documentclass[runningheads]{llncs}
\usepackage{graphicx}
\usepackage{amsmath,amssymb} 
\usepackage{color}
\usepackage[width=122mm,left=30mm,paperwidth=182mm,height=193mm,top=30mm,paperheight=253mm]{geometry}
\usepackage{hyphenat}
\usepackage{enumitem}
\usepackage{booktabs}
\usepackage[pagebackref=false,breaklinks=true,letterpaper=true,colorlinks,bookmarks=false]{hyperref}
\usepackage[Alg.]{algorithm}
\usepackage[algo2e,noline,linesnumbered]{algorithm2e}
\usepackage{array}
\usepackage{placeins}
\usepackage[some,top]{background}
\SetBgColor{black}
\SetBgVshift{-20pt}
\SetBgContents{\text{\underline{A conference version of this paper is accepted at ECCV'16 for oral presentation}}}

\newcolumntype{C}[1]{>{\centering\arraybackslash}p{#1}}
\newcommand{\subscript}[2]{$\mathbf{#1 #2}$}

\begin{document}
\pagestyle{headings}
\mainmatter

\title{Top-down Neural Attention by Excitation Backprop} 

\titlerunning{Top-down Neural Attention by Excitation Backprop}

\authorrunning{J.\ Zhang \emph{et al.}}

\author{Jianming Zhang\inst{1,2}, Zhe Lin\inst{2}, Jonathan Brandt\inst{2}, Xiaohui Shen\inst{2}, Stan Sclaroff\inst{1}}
\institute{Boston University\\
\email{\{jmzhang,sclaroff\}@bu.edu}
\and
Adobe Research\\
\email{\{zlin,jbrandt,xshen\}@adobe.com}
}

\maketitle

\begin{abstract}
We aim to model the top-down attention of a Convolutional Neural Network (CNN) classifier for generating task-specific attention maps. Inspired by a top-down human visual attention model, we propose a new backpropagation scheme, called Excitation Backprop, to pass along top-down signals downwards in the network hierarchy via a probabilistic Winner-Take-All process. Furthermore, we introduce the concept of contrastive attention to make the top-down attention maps more discriminative. In experiments, we demonstrate the accuracy and generalizability of our method in weakly supervised localization tasks on the MS COCO, PASCAL VOC07 and ImageNet datasets. The usefulness of our method is further validated in the text-to-region association task. On the Flickr30k Entities dataset, we achieve promising performance in phrase localization by leveraging the top-down attention of a CNN model that has been trained on weakly labeled web images.
\end{abstract}
\BgThispage
\section{Introduction}

Top-down task-driven attention is an important mechanism for efficient visual search.
Various top-down attention models have been proposed, \emph{e.g.}\ \cite{koch1987shifts,anderson1987shifter,tsotsos1995modeling,wolfe1994guided}. Among them, the Selective Tuning attention model \cite{tsotsos1995modeling} provides a biologically plausible formulation. Assuming a pyramidal neural network for visual processing, the Selective Tuning model is composed of a bottom-up sweep of the network to process input stimuli, and a top-down Winner-Take-ALL (WTA) process to localize the most relevant neurons in the network for a given top-down signal.

Inspired by the Selective Tuning model, we propose a top-down attention formulation for modern CNN classifiers. Instead of the deterministic WTA process used in \cite{tsotsos1995modeling}, which can only generate binary attention maps, we formulate the top-down attention of a CNN classifier as a \emph{probabilistic} WTA process.

The probabilistic WTA formulation is realized by a novel backpropagation scheme, called \emph{Excitation Backprop}, which integrates both top-down and bottom-up information to compute the winning probability of each neuron efficiently. Interpretable attention maps can be generated by Excitation Backprop at intermediate convolutional layers, thus avoiding the need to perform a complete backward sweep. We further introduce the concept of contrastive top-down attention, which captures the differential effect between a pair of contrastive top-down signals. The contrastive top-down attention can significantly improve the discriminativeness of the generated attention maps.

\begin{figure}[t]
  \centering
  \includegraphics[width=1\linewidth]{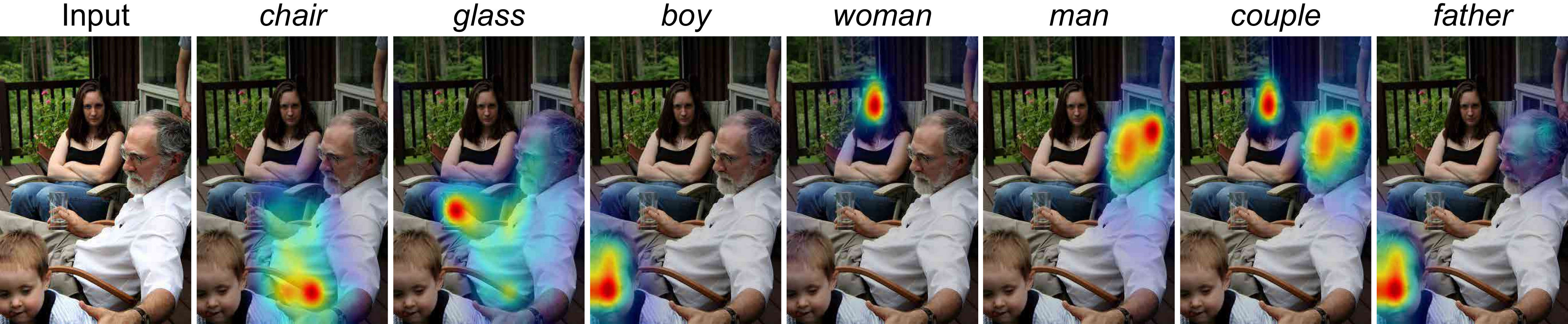}\\
   \caption{A CNN classifier's top-down attention maps generated by our Excitation Backprop can localize common object categories, \emph{e.g.} \texttt{chair} and \texttt{glass}, as well as fine-grained categories like \texttt{boy}, \texttt{man} and \texttt{woman} in this example image, which is resized to $224\times 224$ for our method. The classifier used in this example is trained to predict $\sim$18K tags using only weakly labeled web images. Visualizing the classifier's top-down attention can also help interpret what has been learned by the classifier. For \texttt{couple}, we can tell that our classifier uses the two adults in the image as the evidence, while for \texttt{father}, it mostly concentrates on the child. This indicates that the classifier's understanding of \texttt{father} may strongly relate to the presence of a child.}\label{fig:teaser}
\end{figure}

In experiments, our method achieves superior weakly supervised localization performance \emph{vs.}\ \cite{simonyan2013deep,zeiler2014visualizing,cao2015look,zhou2015learning,bach2015pixel} on challenging datasets such as PASCAL VOC \cite{Everingham10} and MS COCO \cite{lin2014microsoft}.
We further explore the scalability of our method for localizing a large number of visual concepts.
For this purpose, we train a CNN tag classifier to predict $\sim$18K tags using 6M weakly labeled web images.
By leveraging our top-down attention model, our image tag classifier can be used to localize a variety of visual concepts. Moreover, our method can also help to understand
what has been learned by our tag classifier. Some examples are shown in Fig.~\ref{fig:teaser}.

The performance of our large-scale tag localization method is evaluated on the challenging Flickr30k Entities dataset \cite{plummer2015flickr30k}. Without using a language model or any localization supervision, our top-down attention based approach achieves competitive phrase-to-region performance \emph{vs.}\ a fully-supervised baseline \cite{plummer2015flickr30k}.

To summarize, the \textbf{main contributions} of this paper are:
\begin{itemize}
  \item a top-down attention model for CNN based on a probabilistic Winner-Take-All process using a novel Excitation Backprop scheme;
  \item 
  a contrastive top-down attention formulation for enhancing the discriminativeness of attention maps; and
  \item a large-scale empirical exploration of weakly supervised text-to-region association by leveraging the top-down neural attention model.
\end{itemize}

\section{Related Work}


There is a rich literature about modeling the top-down influences on selective attention in the human visual system (see \cite{baluch2011mechanisms} for a review). It is hypothesized that top-down factors like knowledge, expectations and behavioral goals can affect the feature and location expectancy in visual processing \cite{wolfe1994guided,treisman1980feature,koch1987shifts,desimone1995neural}, and bias the competition among the neurons \cite{reynolds2009normalization,tsotsos1995modeling,desimone1995neural,desimone1998visual,beck2009top}. Our attention model is related to the Selective Tuning model of \cite{tsotsos1995modeling}, which proposes a biologically inspired attention model using a top-down WTA inference process.

Various methods have been proposed for grounding a CNN classifier's prediction \cite{simonyan2013deep,zeiler2014visualizing,cao2015look,zhou2015learning,zhou2014object,bach2015pixel}.
In \cite{simonyan2013deep,zeiler2014visualizing,springenberg2014striving}, error backpropagation based methods are used for visualizing relevant regions for a predicted class or the activation of a hidden neuron. Recently, a layer-wise relevance backpropagation method is proposed by \cite{bach2015pixel} to provide a pixel-level explanation of CNNs' classification decisions. Cao \emph{et al.}\ \cite{cao2015look} propose a feedback CNN architecture for capturing the top-down attention mechanism that can successfully identify task relevant regions.  In \cite{zhou2014object}, it is shown that replacing fully-connected layers with an average pooling layer can help generate coarse class activation maps that highlight task relevant regions. Unlike these previous methods, our top-down attention model is based on the WTA principle, and has an interpretable probabilistic formulation. Our method is also conceptually simpler than \cite{cao2015look,zhou2014object} as we do not require modifying a network's architecture or additional training. The ultimate goal of our method goes beyond visualization and explanation of a classifier's decision \cite{zeiler2014visualizing,springenberg2014striving,bach2015pixel}, as we aim to maneuver CNNs' top-down attention to generate highly discriminative attention maps for the benefits of localization.

Training CNN models for weak supervised localization has been studied by \cite{oquab2015object,pathak2015constrained,papandreou2015weakly,pinheiro2015image,fang2015captions}. In \cite{oquab2015object,fang2015captions,pinheiro2015image}, a CNN model is transformed into a fully convolutional net to perform efficient sliding window inference, and then Multiple Instance Learning (MIL) is integrated in the training process through various pooling methods over the confidence score map. Due to the large receptive field and stride of the output layer, the resultant score maps only provide very coarse location information. To overcome this issue, a variety of strategies, \emph{e.g.} image re-scaling and shifting, have been proposed to increase the granularity of the score maps \cite{oquab2015object,pinheiro2015image,pinheiro2013recurrent}. Image and object priors are also leveraged to improve the object localization accuracy in \cite{pathak2015constrained,papandreou2015weakly,pinheiro2015image}. Compared with weakly supervised localization, the problem setting of our task is essentially different. We assume a pre-trained deep CNN model is given, which may not use any dedicated training process or model architecture for the purpose of localization. Our focus, instead, is to model the top-down attention mechanism of \emph{generic} CNN models to produce interpretable and useful task-relevant attention maps.

\section{Method}

\subsection{Top-down Neural Attention based on Probabilistic WTA}\label{sec:pWTA}

\begin{figure}[t]
  \centering
  \includegraphics[width=1\linewidth]{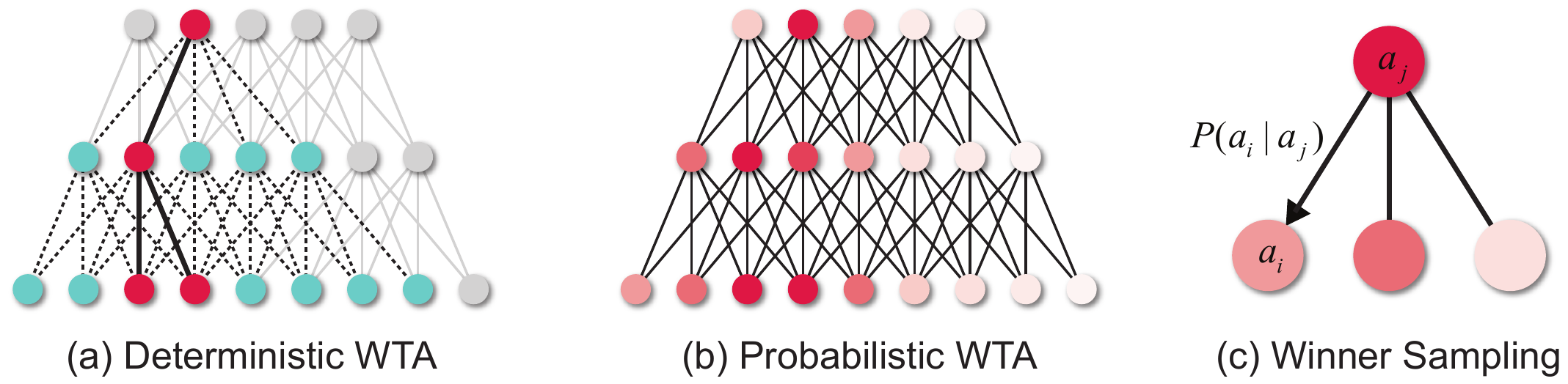}\\
   \caption[Deterministic WTA \emph{vs.}\ our probabilistic WTA]{Deterministic WTA~\cite{tsotsos1995modeling} \emph{vs.}\ our probabilistic WTA for modeling top-down attention. (a)  Given a selected output unit, the red dots denote the winners identified by the top-down layer-wise deterministic WTA scheme in the processing cone, and the cyan ones are inhibited.
  (b) In our probabilistic WTA scheme,  winner neurons are generated by a stochastic sampling process (shown in (c)). The top-down signal is specified by a probability distribution over the output units. The shading of a dot in (b) indicates the its relative likelihood of winning against the other ones in the same layer.}\label{fig:pWTA}
\end{figure}

We consider a generic feedforward neural network model.
The goal of a top-down attention model is to identify the task-relevant neurons in the network.

Given a selected output unit, a deterministic top-down WTA scheme is used in the biologically inspired Selective Tuning model \cite{tsotsos1995modeling} to localize the most relevant neurons in the processing cone (see Fig.~\ref{fig:pWTA} (a)) and generate a binary attention map.
Inspired by the deterministic WTA, we propose a \emph{probabilistic} WTA formulation to model a neural network's top-down attention (Fig.~\ref{fig:pWTA} (b) and (c)), which leverages more information in the network and generates soft attention maps that can capture subtle differences between top-down signals. This is critical to our contrastive attention formulation in Sec.~\ref{sec:contrastive}.

In our formulation, the top-down signal is specified by a prior distribution $P(A_0)$ over the output units, which can model the uncertainty in the top-down control process.
Then the winner neurons are recursively sampled in a top-down fashion based on a conditional winning probability $P(A_t|A_{t-1})$, where $A_t,A_{t-1} \in \mathcal{N}$ denote the selected winner neuron at the current and the previous step respectively, and $\mathcal{N}$ is the overall neuron set. We formulate the top-down relevance of each neuron as its probability of being selected as a winner in this process. Formally, given a neuron $a_j \in \mathcal{N}$ (note that $a_j$ denotes a specific neuron and $A_t$ denotes a variable over the neurons), we would like to compute its \emph{Marginal Winning Probability} (MWP) $P(a_j)$. The MWP $P(a_j)$ can be factorized as
\begin{equation}\label{eqn:pWTA}
  P(a_j) = \sum_{a_i \in \mathcal{P}_j} P(a_j|a_i)P(a_i),
\end{equation}
where $\mathcal{P}_j$ is the parent node set of $a_j$ (in top-down order). As Eqn.~\ref{eqn:pWTA} indicates, given $P(a_j|a_i)$, $P(a_j)$ is a function of the marginal winning probability of the parent nodes in the preceding layers. It follows that $P(a_j)$ can be computed in a top-down layer-wise fashion.

Our formulation is equivalent to an absorbing Markov chain process \cite{kemeny1960finite}. A Markov Chain is an absorbing chain if 1) there is at least one absorbing state and 2) it is possible to go from any state to at least one absorbing state in a finite number of steps. Any walk will eventually end at one of the absorbing states. Non-absorbing states are called Transient States. For an absorbing Markov Chain, the canonical form of the transition matrix $P$ can be represented by
\begin{equation}
\arraycolsep=10pt
P = \left[\begin{array}{cc}
    Q          & R \\
    \mathbf{0} & I_r\\
    \end{array}\right],
\end{equation}
where the entry $p_{ij}$ is the the transition probability from state $i$ to $j$. Each row sums up to one and $I_r$ is an $r\times r$ matrix corresponding to the $r$ absorbing states. In our formulation, each random walk starts from an output neuron and ends at some absorbing node of the bottom layer in the network; and $p_{ij}:= P(a_j|a_i)$ is the transition probability.

The fundamental matrix of the absorbing Markov chain process is
\begin{equation}
 N = \sum_{k=0}^\infty Q^k = (I_t - Q)^{-1},
\end{equation}
The $(i, j)$ entry of $N$ can be interpreted as the the expected number of visits to node $j$, given that the walker starts at $i$. In our formulation, the MWP $P(a_j)$ can then be interpreted as the expected number of visits when a walker starts from a random node of the output layer according to $P(A_0)$. This expected number of visits can be computed by a simple matrix multiplication using the fundamental matrix of the absorbing Markov chain. In this light, the MWP $P(a_i)$ is a linear function of the the top-down signal $P(A_0)$, which will be shown to be convenient later (see Sec.~\ref{sec:contrastive}). In practice, our Excitation Backprop does the computation in a layer-wise fashion, without the need to explicitly construct the fundamental matrix. This layer-wise propagation is possible due to the acyclic nature of the feedforward network.

\subsection{Excitation Backprop}

In this section, we propose the Excitation Backprop method to realize the probabilistic WTA formulation for modern CNN models.

A modern CNN model \cite{krizhevsky2012imagenet,simonyan2014very,szegedy2015going} is mostly composed of a basic type of neuron $a_i$, whose response is computed by $\widehat{a}_i =\varphi(\sum_j w_{ji}\widehat{a}_j + b_j)$. Here $w_{ji}$ is the weight, $\widehat{a}_j$ is the input, $b_j$ is the bias and $\varphi$ is the nonlinear activation function. We call this type of neuron an \emph{Activation Neuron}. We have the following assumptions about the activation neurons.
\setlist[enumerate,1]{leftmargin=0.8cm}
\begin{enumerate}[noitemsep,label=\subscript{A}{\arabic*}.]
  \item The response of the activation neuron is non-negative.
  \item An activation neuron is tuned to detect certain visual features. Its response is positively correlated to its confidence of the detection.
\end{enumerate}

\textbf{A1} holds for a majority of the modern CNN models, as they adopt the Rectified Linear Unit (ReLU) as the activation function. \textbf{A2} has been empirically verified by many recent works \cite{zhou2014object,zeiler2014visualizing,zhou2014learning,yosinski2015understanding}. It is observed that neurons at lower layers detect simple features like edge and color, while neurons at higher layers can detect complex features like objects and body parts.

Between activation neurons, we define a connection to be \emph{excitatory} if its weight is non-negative, and \emph{inhibitory} otherwise. Our Excitation Backprop passes top-down signals through excitatory connections between activation neurons. Formally, let $\mathcal{C}_i$ denote the child node set of $a_i$ (in the top-down order). For each $a_j\in \mathcal{C}_i$, the conditional winning probability $P(a_j|a_i)$ is defined as
\begin{equation}\label{eqn:EB}
  P(a_j|a_i) =
  \begin{cases}
      Z_i\widehat{a}_j w_{ji} & \text{if } w_{ji} \geq 0,\\
      0  & \text{otherwise.}
  \end{cases}
\end{equation}
$Z_i=1/\sum_{j:w_{ji}\geq 0} \widehat{a}_j w_{ji}$ is a normalization factor so that $\sum_{a_j\in \mathcal{C}_i} P(a_j|a_i) = 1$. In the special case when $\sum_{j:w_{ji}\geq 0} \widehat{a}_j w_{ji} = 0$, we define $Z_j$ to be 0. Note that the formulation of $P(a_j|a_i)$ is valid due to \textbf{A1}, since $\widehat{a}_j$ is always non-negative.

Eqn.~\ref{eqn:EB} assumes that if $a_i$ is a winner neuron, the next winner neuron will be sampled among its child node set $\mathcal{C}_i$ based on the connection weight $w_{ji}$ and the input neuron's response $\widehat{a}_j$. The weight $w_{ji}$ captures the top-down feature expectancy, while $\widehat{a}_j$ represents the bottom-up feature strength, as assumed in \textbf{A2}. Due to \textbf{A1}, child neurons of $a_i$ with negative connection weights always have an inhibitory effect on $a_i$, and thus are excluded from the competition.

\begin{figure}[t]
  \centering
  \includegraphics[width=1\linewidth]{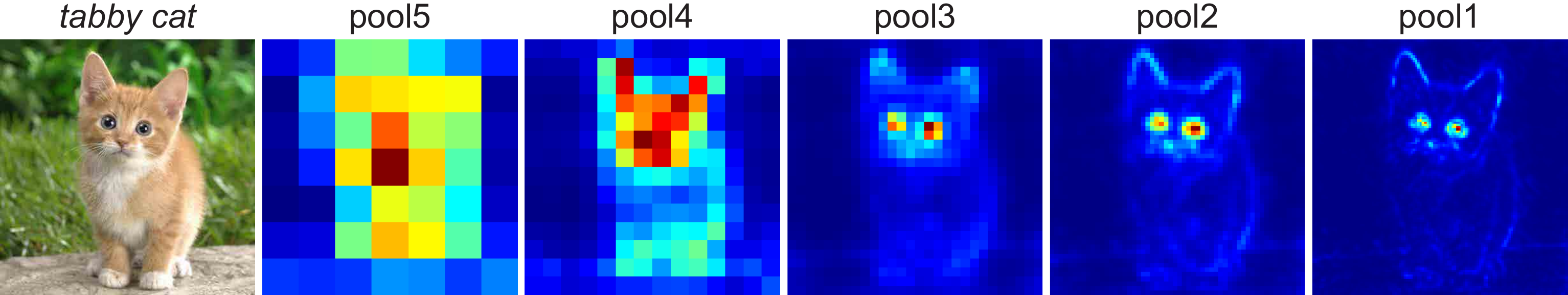}\\
  \caption[Example Marginal Winning Probability (MWP) maps]{Example Marginal Winning Probability (MWP) maps computed via Excitation Backprop from different layers of the public \texttt{VGG16} model \cite{simonyan2014very} trained on ImageNet. The input image is shown on the right. The MWP maps are generated for the category \texttt{tabby cat}. Neurons at higher-level layers have larger receptive fields and strides. Thus, they can capture larger areas but with lower spatial accuracy. Neurons at lower layers tend to more precisely localize features at smaller scale.}\label{fig:EBvsEB}
\end{figure}

Eqn.~\ref{eqn:EB} recursively propagates the top-down signal layer by layer, and we can compute attention maps from any intermediate convolutional layer. For our method, we simply take the sum across channels to generate a marginal winning probability (MWP) map as our attention map, which is a 2D probability histogram. Fig.~\ref{fig:EBvsEB} shows some example MWP maps generated using the pre-trained \texttt{VGG16} model \cite{simonyan2014very}. Neurons at higher-level layers have larger receptive fields and strides. Thus, they can capture larger areas but with lower spatial accuracy. Neurons at lower layers tend to more precisely localize features at smaller scales.

For a class of activation functions that is lower bounded, \emph{e.g.} the sigmoid function, $\tanh$ function and the Exponential Linear Unit (ELU) function \cite{clevert2015fast}, we can slightly modify our formulation of Excitation Backprop. Suppose $\lambda$ is the minimum value in the range of the activation function. The modified formulation corresponding to Eqn.~\ref{eqn:EB} in our paper is
\begin{equation}\label{eqn:EB2}
  P(a_j|a_i) =
  \begin{cases}
      Z_i(\widehat{a}_j + \lambda) w_{ji} & \text{if } w_{ji} \geq 0,\\
      0  & \text{otherwise.}
  \end{cases}
\end{equation}
Because $\widehat{a}_j + \lambda \geq 0$ , our probability formulation still holds.

\subsection{Contrastive Top-down Attention}\label{sec:contrastive}

Since the MWP is a linear function of the top-down signal (see Sec.~\ref{sec:pWTA}), we can compute any linear combination of MWP maps for an image by a single backward pass. All we need to do is linearly combine the top-down signal vectors
at the top layer before performing the Excitation Backprop. In this section, we take advantage of this property to generate highly discriminative top-down attention maps by passing down pairs of contrastive signals.

\begin{figure}[t]
  \centering
  \includegraphics[width=1\linewidth]{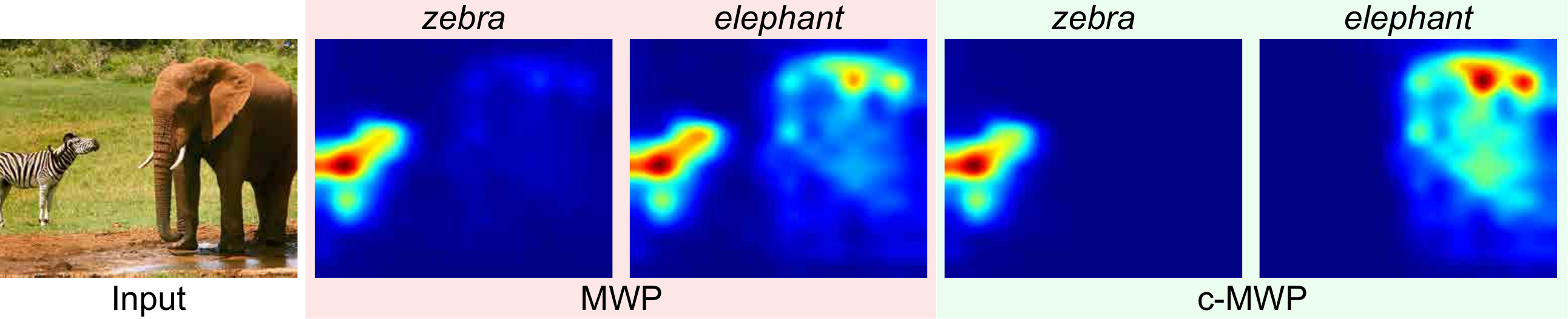}\\
  \caption[Marginal Winning Probability \emph{vs.}\ contrastive MWP]{Marginal Winning Probability (MWP) \emph{vs.}\ contrastive MWP (c-MWP). The input image is resized to 224$\times$224, and we use \texttt{GoogleNet} pretrained on ImageNet to generate the MWP maps and c-MWP maps for \texttt{zebra} and \texttt{elephant}. The MWP map for \texttt{elephant} does not successfully suppress the zebra. In contrast, by cancelling out common winner neurons for \texttt{elephant} and \texttt{non-elephant}, the c-MWP map more effectively highlights the elephant.}\label{fig:contrastive}
\end{figure}

For each output unit $o_i$, we virtually construct a dual unit $\bar{o}_i$, whose input weights are the negation of those of $o_i$. For example, if an output unit corresponds to an \texttt{elephant} classifier, then its dual unit will correspond to a \texttt{non-elephant} classifier. Subtracting the MWP map for \texttt{non-elephant} from the one for \texttt{elephant} will cancel out common winner neurons and amplify the discriminative neurons for \texttt{elephant}. We call the resulting map a \emph{contrastive} MWP map, which can be computed by a single backward pass. Fig.~\ref{fig:contrastive} shows some examples.

Formally, let $W_1$ be the weights of the top layer, and $P_1$ be the corresponding transition matrix whose entries are the conditional probabilities defined by Eqn.~\ref{eqn:EB}. Suppose the number of the neurons at the top is $m$ and at the next lower layer is $n$, and $P_1$ is a $m\times n$ matrix.

The weights of the contrastive output units are the negation of the original weights at the top layer, namely $-W_1$. Let $\bar{P}_1$ denote the corresponding transition matrix. Regarding $\bar{P}_1$, the entries that are positive were previously thresholded in $P_1$ according to Eqn.~\ref{eqn:EB} and \emph{vise versa}. For example, $p_{ij}>0$ in $P_1$ indicates $\bar{p}_{ij}=0$ in $\bar{P}_1$.

The MWP of a target layer, say the n-th layer from the top, is formulated as
\begin{equation}
C = P_0 \cdot P_1 \cdot P_2 \cdot \ldots \cdot P_{n-1},
\end{equation}
and the dual MWP for the contrastive output units is
\begin{equation}
\bar{C} = P_0 \cdot \bar{P}_1\cdot P_2 \cdot \ldots \cdot P_{n-1},
\end{equation}
where $P_0$ is the input top-down signal in the form of a horizontal vector. The resultant contrastive MWP is formulated as
\begin{equation}
C-\bar{C} = P_0 \cdot (P_1 - \bar{P}_1) \cdot P_2 \cdot \ldots \cdot P_{n-1}.
\end{equation}

In practice, we compute $P_0\cdot P_1$ and $P_0 \cdot \bar{P}_1$ respectively by Excitation Backprop. Then, we do the subtraction and propagate the contrastive signals $P_0\cdot(P_1 - \bar{P}_1)$ downwards by Excitation Backprop again. Moreover, we truncate the contrastive MWP map at zero so that only positive parts are kept. Our probabilistic formulation ensures that there are always some positive parts on the contrastive MWP map, unless the MWP map and its dual are identical.

\subsection{Implementation of Excitation Backprop}

We implement Excitation Backprop in Caffe \cite{jia2014caffe} (available at our project website\footnote{\url{http://www.cs.bu.edu/groups/ivc/excitation-backprop}}). In the following, we describe the implementation of Excitation Backprop for common layers in modern CNNs.

\textbf{ReLU Layer}: The ReLU layer serves as a neuron-wise gating function. The top-down signals remain the same after a ReLU layer in Excitation Backprop, since each ReLU neuron only has a single child node. Neurons with zero activation values will not be selected as winning neurons due to Eqn.~\ref{eqn:EB} in our paper. Thus, the propagation method for ReLU is the same as error backpropagation.

\textbf{Max Pooling Layer}: Again, as in error backpropagation, signals are copied to the lower layer through the pooling mask, because each pooled neuron has only a single child node. Therefore, the propagation method for the Max Pooling layer is the same as error backpropagation, too.

\textbf{Convolutional, fully-connected and Average Pooling Layers}: Convolutional, fully-connected and average pooling layers can be regarded as the same type of layers that perform an affine transform of the response values of the bottom (input) neurons. The implementation of Excitation Backprop for these layers exactly follows Eqns.~\ref{eqn:pWTA} and \ref{eqn:EB} in our paper. Let $p_{ij}:=P(a_j|a_i)$ and $p_j:=P(a_j)$ for convenience, and we have:
\begin{align}
    p_j &= \sum_{i\in \mathcal{P}_j} p_{ij} p_i \nonumber\\
        &= \sum_{i\in \mathcal{P}_j} Z_i\widehat{a}_j w^+_{ji}p_i \nonumber\\
        &= \widehat{a}_j\sum_{i\in \mathcal{P}_j} w^+_{ji}p_i\frac{1}{\sum_{k\in \mathcal{C}_i}w^+_{ki}\widehat{a}_k},
\end{align}
where $w^+_{ji}=\max\{w_{ji},0\}$, $\mathcal{P}_i$ is the parent node set of $a_j$ and $\mathcal{C}_i$ is the child node set of $a_i$ (in top-down order). The computation of all $p_j$ in a layer can be performed by matrix operations:
\begin{equation}\label{eqn:EB1}
    P_{n} = A_{n}\odot\left( W^+\left(P_{n-1} \oslash({W^+}^T A_{n}) \right) \right),
\end{equation}
where $P_{n-1}$ and $P_{n}$ denote the Marginal Winning Probability (MWP) for the top neurons and the bottom neurons of the layer respectively, and $W^+=\left[w^+_{ij}\right]_{d_1,d_2}$ is a $d_1\times d_2$ weight matrix representing the excitatory connection weight of the layer. $d_1$ ($d_2$) equals the number of the bottom (top) neurons. $A_{n}$ is the response value of the bottom neurons. Note that we assume that the response values of the bottom neurons are non-negative; thus, we do not propagate the top-down signals to the mean-subtracted pixel layer, which may contain negative pixel values. Moreover, $\odot$ and $\oslash$ are the element-wise multiplication and division respectively. Alg.~\ref{alg:EB} summarizes the steps of Excitation Backprop for the convolutional layer.

\begin{algorithm}[t]
\SetKwInOut{Input}{input}\SetKwInOut{Output}{output}
\SetKwInOut{Aux}{auxiliaries}
\SetAlgoVlined
\caption{Excitation Backprop for the Convolutional Layer}
\label{alg:EB}
\Input{$A_{n}$: bottom activation responses;\\$W$: weight parameters;\\$P_{n-1}$: top MWP}
\Output{$P_{n}$: bottom MWP}
\BlankLine
compute $W^+$ by thresholding $W$ at zero

compute $X={W^+}^T A_{n}$ in Eqn.~\ref{eqn:EB1} (a \texttt{forward} layer operation in Caffe)

compute $Y=P_{n-1} \oslash X$ by element-wise division

compute $Z = W^+ Y$  (a \texttt{backward} layer operation in Caffe)

compute $P_{n} = A_{n}\odot Z$ by element-wise multiplication
\end{algorithm}

\textbf{Local Response Normalization (LRN) Layer.} Special care should be taken for the LRN layer, whose response is computed element-wise as $\widehat{m}_i = s_i \widehat{a}_i$, where $s_i$ is a positively valued scaling factor computed based on the neighboring neurons' response. The LRN layer locally normalizes each neuron's response. In Excitation Backprop, we just ignore the normalization factor, and thus each neuron $m_i$ of the LRN layer has only one child node $a_i$ in the top-down propagation process. As a result, the top-down signals remain the same when passing through the LRN layer.

\section{Experiments}

\subsection{The Pointing Game}

The goal of this section is to evaluate the \emph{discriminativeness} of different top-down attention maps for localizing target objects in crowded visual scenes.

\textbf{Evaluation setting.} Given a pre-trained CNN classifier, we test different methods in generating a top-down attention map for a target object category present in an image. Ground truth object labels are used to cue the method. We extract the maximum point on the top-down attention map. A hit is counted if the maximum point lies on one of the annotated instances of the cued object category, otherwise a miss is counted. We measure the localization accuracy by $Acc = \frac{\# Hits}{\# Hits + \#Misses}$ for each object category. The overall performance is measured by the mean accuracy across different categories.

We call this the \emph{Pointing Game}, as it asks the CNN model to point at an object of designated category in the image. The pointing game does not require highlighting the full extent of an object, and it does not account for the CNN model's classification accuracy. Therefore, it purely compares the \emph{spatial selectiveness} of the top-down attention maps. Moreover, the pointing game only involves minimum post-processing of the attention maps, so it can evaluate different types of attention maps more fairly.

\textbf{Datasets.} We use the test split of the PASCAL VOC07 dataset \cite{Everingham10} (4952 images) and the validation split of the MS COCO dataset \cite{lin2014microsoft} (40137 images). In particular, COCO contains 80 object categories, and many of its images have multiple object categories, making even the simple Pointing Game rather challenging. To evaluate success in the Pointing Game, we use the groundtruth bounding boxes for VOC07 and the provided segmentation masks for COCO.

\textbf{CNN classifiers.} We consider three popular CNN architectures: \texttt{CNN-S} \cite{Chatfield14} (an improved version of AlexNet \cite{krizhevsky2012imagenet}), \texttt{VGG16} \cite{simonyan2014very}, and \texttt{GoogleNet} \cite{szegedy2015going}. These models vary a lot in depth and structure. We download these models from the Caffe Model Zoo website\footnote{\url{https://github.com/BVLC/caffe/wiki/Model-Zoo}}. These models are pre-trained on ImageNet \cite{ILSVRC15}. For both VOC07 and COCO, we use the training split to fine-tune each model. We follow the basic training procedure for image classification. Only the output layer is fine-tuned using the multi-label cross-entropy loss for simplicity, since the classification accuracy is not our focus. Images are padded to square shape by mirror padding and up-sampled to 256$\times$256. Random flipping and cropping are used for data augmentation. No multi-scale training \cite{oquab2015object} is used. We fix the learning rate to be $0.01$ for all the architectures and optimize the parameters using SGD. The training batch size is set as 64, 32 and 64 for VGGS, VGG16 and GoogleNet respectively. We stop the training when the training error plateaus.


\textbf{Test methods.} We compare Excitation Backprop (MWP and c-MWP) with the following methods: (Grad) the error backprogation method \cite{simonyan2013deep}, (Deconv) the deconvolution method originally designed for internal neuron visualization \cite{zeiler2014visualizing}, (LRP) layer-wise relevance propagation \cite{bach2015pixel}, and (CAM) the class activation map method \cite{zhou2015learning}.
We implement Grad, Deconv and CAM in Caffe. For Deconv, we use an improved version proposed in \cite{springenberg2014striving}, which generates better maps than the original version \cite{zeiler2014visualizing}. For Grad and Deconv, we follow \cite{simonyan2013deep} to use the maximum absolute value across color channels to generate the final attention map. Taking the mean instead of maximum will degrade their performance. For LRP, we use the software provided by the authors, which only supports CPU computation. For \texttt{VGG16}, this software can take 30s to generate an attention map on an Intel Xeon 2.90GHz$\times$6 machine\footnote{On COCO, we need to compute about 116K attention maps, which leads to over 950 hours of computation on a single machine for LRP using \texttt{VGG16}.}. Due to limited computational resources, we only evaluate LRP for \texttt{CNN-S} and \texttt{GoogleNet}.

Note that CAM is only applicable to certain architectures like \texttt{GoogleNet}, which do not have fully connected layers. At test time, it acts like a fully convolutional model to perform dense sliding window evaluation \cite{oquab2015object,sermanet2013overfeat}. Therefore, the comparison with CAM encompasses the comparison with the dense evaluation approach for weakly supervised localization \cite{oquab2015object}.

To generate the full attention maps for images of arbitrary aspect ratios, we convert each testing CNN classifier to a fully convolutional architecture as in \cite{oquab2015object}. All the compared methods can be easily extended to fully convolutional models. In particular, for Excitation Backprop, Grad and Deconv, the output confidence map of the target category is used as the top-down signal to capture the spatial weighting. However, all input images are resized to 224 in the smaller dimension, and no multi-scale processing is used.

For different CNN classifiers, we empirically select different layers to compute our attention maps based on a held-out set. We use the conv5 layer for \texttt{CNN-S}, pool4 for \texttt{VGG16} and pool2 for \texttt{GoogleNet}. We use bicubic interpolation to upsample the generated attention maps. The effect of the layer selection will be analysed below. For Grad, Deconv and LRP we blur their maps by a Gaussian kernel with $\sigma = 0.02\cdot\max\{W,H\}$, which slightly improves their performance since their maps tend to be sparse and noisy at the pixel level. In the evaluation, we expand the groundtruth region by a tolerance margin of 15 pixels, so that the attention maps produced by CAM, which are only 7 pixels in the shortest dimension, can be more fairly compared.

\begin{table}[t]
  \centering
  \caption[Mean accuracy in the Pointing Game]{Mean accuracy (\%) in the Pointing Game. For each method, we report two scores for the overall test set and a difficult subset respectively. \texttt{Center} is the baseline that points at image center. The second best score of each column is underlined.}\label{tab:pointgame}
  \scriptsize
  \begin{tabular}{rC{1.5cm}C{1.5cm}C{1.5cm}C{0.2cm}C{1.5cm}C{1.5cm}C{1.5cm}}
    \toprule
                                       & \multicolumn{3}{c}{VOC07 Test (All/Diff.)} & & \multicolumn{3}{c}{COCO Val. (All/Diff.)} \\\cmidrule{2-4}\cmidrule{6-8}
                                       & CNN-S             & VGG16            & GoogleNet       && CNN-S            & VGG16           & GoogleNet    \\\cmidrule{1-1}
    Center                             & 69.5/42.6        & 69.5/42.6        & 69.5/42.6       && 27.7/19.4            & 27.7/19.4            & 27.7/19.4         \\\cmidrule{1-1}
    Grad \cite{simonyan2013deep}       & \underline{78.6}/\underline{59.8} & 76.0/\underline{56.8} & 79.3/61.4   && \underline{38.7}/\underline{30.1}& 37.1/30.7  & 42.6/36.3         \\
    Deconv \cite{zeiler2014visualizing}& 73.1/45.9        & 75.5/52.8        & 74.3/49.4       && 36.4/28.4      & 38.6/30.8 & 35.7/27.9         \\
    LRP \cite{bach2015pixel}           & 68.1/41.3        & -                & 72.8/50.2            && 32.5/24.0               &  -              & 40.2/32.7            \\
    CAM \cite{zhou2015learning}        & -                & -                & \underline{80.8}/\underline{61.9} && -               &  -              & 41.6/35.0        \\\cmidrule{1-1}
    MWP                                & 73.7/52.9             & \underline{76.9}/55.1             & 79.3/60.4            && 35.0/27.7   &  \underline{39.5}/\underline{32.5}  & \underline{43.6}/\underline{37.1}\\
    c-MWP                              & \textbf{78.7}/\textbf{61.7}    & \textbf{80.0}/\textbf{66.8}    & \textbf{85.1}/\textbf{72.3}   && \textbf{43.0}/\textbf{37.0}   &  \textbf{49.6}/\textbf{44.2}  & \textbf{53.8}/\textbf{48.3}\\

    \bottomrule
  \end{tabular}
\end{table}

\begin{figure}[t!]
  \centering
  \includegraphics[width=1\linewidth]{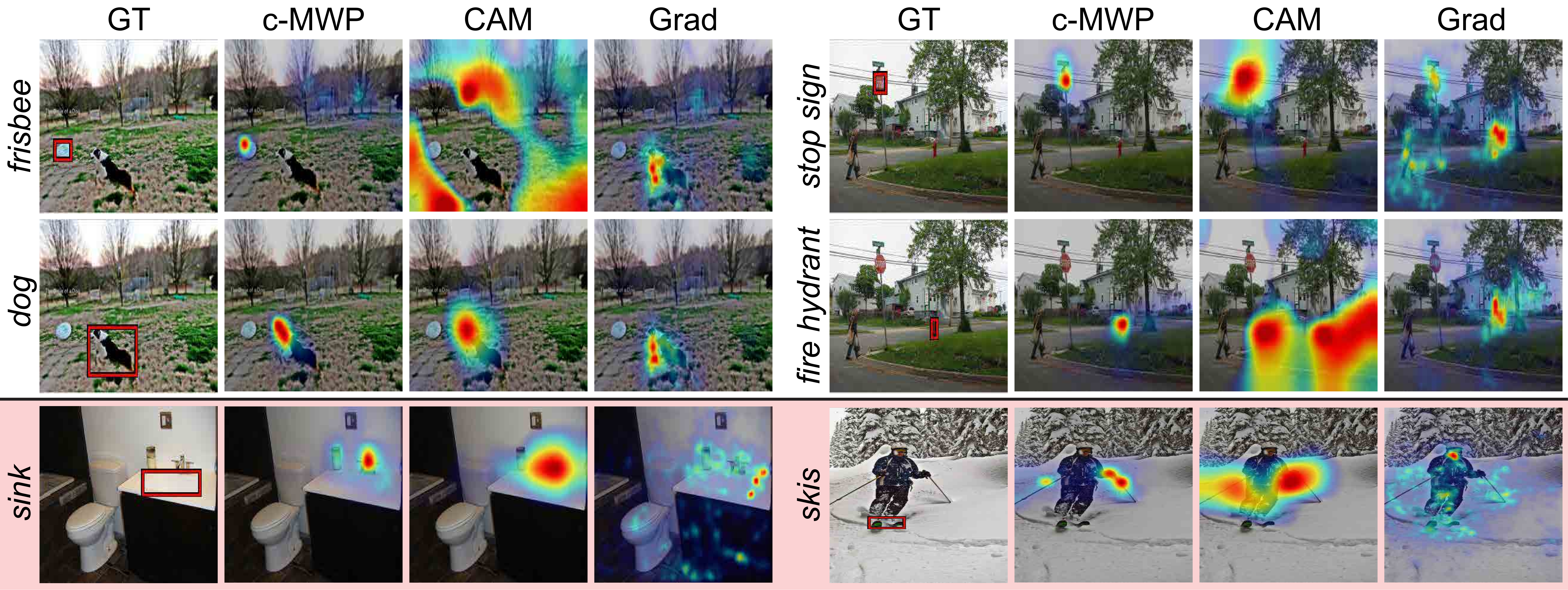}\\
  \caption[Example attention maps using GoogleNet]{Example attention maps using \texttt{GoogleNet}. For visualization, the maps are superimposed on the images after some postprocessing (slight blur for Grad and thresholding for CAM). (\emph{Top two rows}) Our c-MWP is very discriminative and can often localize challengingly small objects like \texttt{frisbee}, \texttt{stop sign} and \texttt{fire hydrant}. (\emph{Bottom row}) Two typical failure cases of top-down neural attention are shown. Since \texttt{faucet} often co-occurs with \texttt{sink}, the CNN's attention falsely focuses on the faucet in the image. It is the same case for \texttt{ski poles} and \texttt{skis}.}\label{fig:example}
\end{figure}


\textbf{Results.} The results are reported in Table \ref{tab:pointgame}. As the pointing game is trivial for images with large dominant objects, we also report the performance on a difficult subset of images for each category. The difficult set includes images that meet two criteria: 1) the total area of bounding boxes (or segments in COCO) of the testing category is smaller than $1/4$ the size of the image and 2) there is at least one other distracter category in the image.

Our c-MWP consistently outperforms the other methods on both VOC07 and COCO across different CNN models. c-MWP is also substantially better than MWP, which validates the idea of contrastive attention. \texttt{GoogleNet} provides the best localization performance for different methods, which is also observed by \cite{cao2015look,zhou2015learning}. Using \texttt{GoogleNet}, our c-MWP outperforms the second best method by about 10 percentage points on the difficult sets of VOC07 and COCO. In particular, we find that our c-MWP gives the best performance in 69/80 object categories of COCO, especially for small objects like \texttt{remote}, \texttt{tie} and \texttt{baseball bat}.

Example attention maps are shown in Fig.~\ref{fig:example}. As we can see, our c-MWP maps can accurately localize the cued objects in rather challenging scenes.

\begin{figure}[t]
  \centering
  \includegraphics[width=1\linewidth]{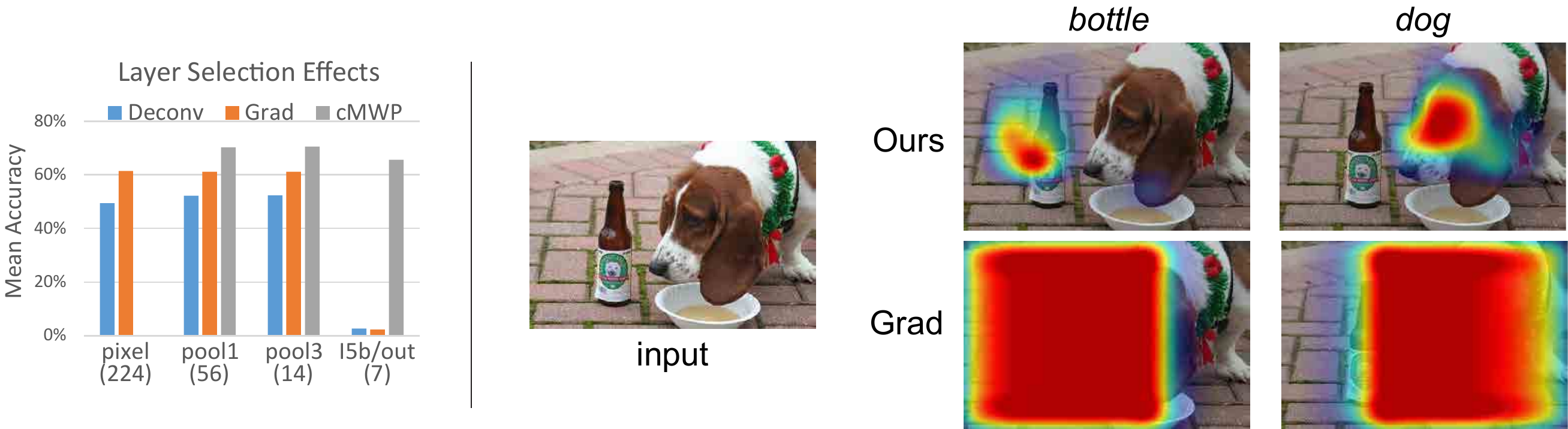}\\
  \caption[Effects of layer selection on VOC07 difficult set]{Effects of layer selection on VOC07 \emph{difficult} set. (\emph{Left}) For Grad, Deconv and our c-MWP, we compare their attention maps from three different layers in the \texttt{GoogleNet}. At I5b/out, Grad and Deconv fail to generate meaningful attention maps, while our method can still achieve reasonable accuracy. (\emph{Right}) We show example attention maps by our c-MWP and Grad from the I5b/out layer.}\label{fig:LS}
\end{figure}

\textbf{Layer selection effects.} We use \texttt{GoogleNet} to analyze the effects of layer selection. For a comparison, we also report the performance of Grad and Deconv by taking the maximum gradient magnitude across feature map channels in the intermediate layers. Results are reported in Fig.~\ref{fig:LS}. We choose three intermediate layers in \texttt{GoogleNet}: pool1, pool3 and Inception\_5b/output (I5b/out), whose spatial resolutions are 56, 14 and 7 in the shortest dimension respectively. Performance does not vary much across all methods at the chosen layers except I5b/out. Our c-MWP only gets a slight decrease in accuracy (mainly due to the map's low spatial resolution), while Grad and Deconv do not generate meaningful attention maps (see Fig.~\ref{fig:LS}). This is because the attention maps of Grad and Deconv at I5b/out are not conditioned on the activation values of I5b/out, and thus fail to leverage the spatial information captured by I5b/out.

\textbf{Analysis of contrastive top-down attention.} The proposed contrastive attention is conceptually simple, which basically subtracts one attention map from its dual using the virtual contrastive output unit. We test this idea for Grad, Deconv and CAM and the performance is reported in Table \ref{tab:module}. For Grad, the gradient magnitude map is identical to its dual since the gradients of the dual map are just the negation of the reference map. As a result, the subtraction gives a zero map. For CAM, the performance remains the same because the dual map is again a negation of the reference attention map and the maximum point will not be changed by the subtraction. However, the proposed contrastive attention works for Deconv, when the attention map and its dual are L1-normalized before subtraction. Deconv shares a similar spirit of our method as it discards negative/inhibitatory signals by thresholding at ReLU layers, but it also introduces non-linearity in the propagation process. Therefore, it requires two backward passes and proper normalization, while our method can directly propagate the contrastive signal via a single pass and achieves better performance.

\begin{table}[t]
  \centering
  \caption[Analysis of contrastive attention on VOC07 difficult set using GoogleNet]{Analysis of contrastive attention on VOC07 \emph{difficult} set using \texttt{GoogleNet}. We evaluate two variants of Excitation Backprop for the contrastive attention map computation compared with our full model. We also test the contrastive attention idea for Grad, Deconv and CAM and their original scores are shown in brackets. See text for details.}\label{tab:module}
  \setlength{\tabcolsep}{5pt}
  \scriptsize
  \begin{tabular}{cccccccc}
    \toprule
                   &\multicolumn{3}{c}{Excitation Backprop} &&\multicolumn{3}{c}{Other Methods} \\\cmidrule{2-4}\cmidrule{6-8}
                   &full &  post-norm & w/o norm &&c-Grad &c-Deconv & c-CAM\\
    Mean Acc. (\%) &\textbf{70.6} & 58.1        & 41.6      &&N.A.     & 67.7 (49.4)  & 61.9 (61.9) \\
    \bottomrule
  \end{tabular}
\end{table}

Our probabilistic WTA formulation produces well-normalized attention maps that enable direct subtraction. We report the performance of two variants of our method in Table \ref{tab:module}. We remove the normalization factor $Z_i$ in Eqn.~\ref{eqn:EB} and pass down the contrastive signal. This leads to a significant degradation in performance (w/o norm). Then we compute the attention map and its dual separately and do the subtraction after L1-normalization (post-norm). The performance is improved but still substantially lower than our full method. This analysis further confirms the importance of our probabilistic formulation.

\subsection{Localizing Dominant Objects}\label{sec:WSL}

We now turn to a different evaluation setting \cite{cao2015look}. The goal of this setting is bounding box (bbox) localization of dominant objects in the image.

\textbf{Dataset and evaluation.} We follow the protocol of Feedback Net \cite{cao2015look} for a fair comparison. The test is performed on the ImageNet Val.\ set ($\sim$50K images), where each image has a label representing the category of dominant objects in it. The label is given, so the evaluation is based on the localization error rate with an IOU threshold at 0.5. Images are resized to 224$\times$224.

As in \cite{cao2015look}, simple thresholding is used to extract a bbox from an attention map. We set the threshold $\tau = \alpha \mu_I$, where $\mu_I$ is the mean value of the map. Then the tightest bbox covering the white pixels is extracted. The parameter $\alpha$ is optimized in the range $[0:0.5:10]$ for each method on a held out set.

\textbf{Results.} Table \ref{tab:WSL} reports the results based on the same \texttt{GoogleNet} model obtained from Caffe Model Zoo as in \cite{cao2015look}. We find that c-MWP performs poorly, but our MWP obtains competitive results against Feedback and other methods. Compared with Feedback, our method is conceptually much simpler. Feedback requires modification of a CNN's architecture and needs 10-50 iterations of forward-backward passes for computing an attention map.

\begin{table}[t]
  \centering
  \caption[Bounding box localization error on ImagNet Val.\ using GoogleNet]{Bounding box localization error on ImagNet Val.\ using \texttt{GoogleNet}. $^*$The score of Feedback is from the original paper.}\label{tab:WSL}
  \scriptsize
  \begin{tabular}{ccccc|c|C{1.1cm}C{1.1cm}}
    \bottomrule
               & Grad \cite{simonyan2013deep}                   & Deconv \cite{zeiler2014visualizing}                     & LRP \cite{bach2015pixel}               & CAM  \cite{zhou2015learning}                 & Feedback$^*$ \cite{cao2015look}     & c-MWP & MWP \\\hline
    Opt. $\alpha$     & 5.0  & 4.5  & 1.0  & 1.0  & -                & 0.0 & 1.5\\
    Loc. Error (\%)   & 41.6 & 41.6 & 57.8 & 48.1 & \underline{38.8} & 57.0 & \textbf{38.7} \\
    \toprule
  \end{tabular}
\end{table}

Note that this task favors attention maps that fully cover the \emph{dominant} object in an image. Thus, it is very different from the Pointing Game, which favors discriminativeness instead. Our c-MWP usually only highlights the most discriminative part of an object due to the competition between the contrastive pair of top-down signals.
This experiment highlights the versatility of our method, and the value of the non-contrastive version (MWP) for dominant object localization.



\subsection{Text-to-Region Association}

Text-to-region association in unconstrained images \cite{plummer2015flickr30k} is very challenging compared to the object detection task, due to the lack of fully-annotated datasets and the large number of words/phrases used in the natural language. Moreover, an image region can be referred to by potentially many different words/phrases, which further increases the complexity of the fully-supervised approach.

By leveraging the top-down attention of a CNN image tag classifier, we propose a highly scalable approach to weakly supervised word-to-region association. We train an image tag classifier using $\sim$6M weakly labeled thumbnail images collected from a commercial stock image website\footnote{\url{https://stock.adobe.com}} (Stock6M). Each image is 200-pixels in the longest dimension and comes with about 30-50 user tags. These tags cover a wide range of concepts, including objects, scenes, body parts, attributes, activities, and abstract concepts, but are also very noisy. We picked $\sim$18K most frequent tags for our dictionary. We empirically found that the first few tags of each image are usually more relevant, and consequently use only the first 5 tags of an image in the training.

\textbf{Tag classifier training.} We use the pre-trained \texttt{GoogleNet} model from Caffe Model Zoo, and fine-tune the model using the multi-label cross-entropy objective function for the 18K tags. Images are padded to square shape by mirror padding and upsampled to 256$\times$256. Random flipping and cropping are used for data augmentation. We use SGD with a batch size of 64 and a starting learning rate of 0.01. The learning rate is lowered by a factor of 0.1 when the validation error plateaus. The training process passes through the data for three epochs and takes $\sim$55 hours on an NVIDIA K40c GPU.

\textbf{Dataset and evaluation.} To quantitatively evaluate our top-down attention method and the baselines in text-to-region association, we use the recently proposed Flickr30k Entities (Flickr30k) dataset \cite{plummer2015flickr30k}. Evaluation is performed on the test split of Flickr30k (1000 images), where every image has five sentential descriptions. Each Noun Phrase (NP) in a sentence is manually associated with the bounding box (bbox) regions it refers to in the image. NPs are grouped into eight types (see \cite{plummer2015flickr30k}). Given an NP, the task is to provide a list of scored bboxes, which will be measured by the recall rate (similar to the object proposal metric) or per-group/per-phrase Average Precision (AP) (similar to the object detection metric). We use the evaluation code from \cite{plummer2015flickr30k}.

\begin{table}[t]
  \centering
  \caption[Performance comparison on the Flickr30k Entities dataset]{Performance comparison on the Flickr30k Entities dataset. We report performance for both the whole dataset and a subset of small instances. The R@$N$ refers to the overall recall rate regardless of phrase types. mAP (Group) and mAP (Phrase) should be interpreted differently, because most phrases belong to the group \texttt{people}. CCA$^*$ refers to the precomputed results provided by \cite{plummer2015flickr30k}, while CCA is the results reported in the original paper. MCG\_base is the performance using MCG's original proposal scores. EB is EdgeBoxes \cite{zitnick2014edge}.}\label{tab:Flickr_overall}
  \scriptsize
  \setlength{\tabcolsep}{4pt}
  \begin{tabular}{r|c|ccc|cc}
    \bottomrule
              & opt. $\gamma$ & R@1   & R@5     & R@10     & mAP (Group) & mAP (Phrase) \\\hline
    MCG\_base & --            & 10.7/\,\,7.7  & 30.3/22.4    & 40.5/30.3     & \,\,6.9/\,\,4.5         & 16.8/12.9    \\\hline
    Grad (MCG)      & 0.50           & 24.3/\,\,7.6  & 49.6/32.9    & 59.7/45.8     & 10.2/\,\,3.8        & 28.8/15.6   \\
    Deconv (MCG)    & 0.50           & 21.5/11.3  & 48.4/34.5    & 58.5/46.0     & 10.0/\,\,4.0        & 26.5/16.7   \\
    LRP (MCG)      & 0.50           & 24.3/11.8  & 51.6/36.8    & 61.3/48.5     & 10.3/\,\,4.3        & 28.9/18.1   \\
    CAM (MCG)      & 0.75          & 21.7/\,\,6.5  & 47.1/27.9    & 56.1/39.1     & \,\,7.5/\,\,2.0         & 26.0/11.9   \\\hline
    MWP (MCG)      & 0.50           & \textbf{28.5}/15.0          & 52.7/39.1         & 61.3/49.8          & 11.8/\,5.3           & \textbf{31.1}/20.3 \\
    c-MWP (MCG)     & 0.50          &26.2/\textbf{21.2}  & \textbf{54.3}/\textbf{43.4}    & \textbf{62.2}/\textbf{51.7}     & \textbf{15.2}/\textbf{10.8}        & 30.8/\textbf{24.0}   \\\toprule\bottomrule
    CCA$^*$ \cite{plummer2015flickr30k} (EB)& -- & 25.2/\textbf{21.8}   & \textbf{50.3}/\textbf{41.0}  & 58.1/\textbf{47.3} & 12.8/\textbf{11.5}  & 28.8/\textbf{23.6} \\
    CCA \cite{plummer2015flickr30k} (EB) & -- & 25.3/\,\,\,\,--\,\,\,\,       & --     & \textbf{59.7}/\,\,\,\,--\,\,\,\, & 11.2/\,\,\,\,--\,\,\,\,  & --    \\\hline
    c-MWP (EB) & 0.25          & \textbf{27.0}/18.4  & 49.9/35.2 & 57.7/43.9 & \textbf{13.2}/\,\,8.1 & \textbf{29.4}/20.0 \\
    \toprule
  \end{tabular}
\end{table}
\begin{table}[t]
  \centering
  \caption[Per group recall@5 on the Flickr30k Entities dataset]{Per group recall@5 (\%) on the Flickr30k Entities dataset. The mean scores are computed over different group types, which are different from the overall recall rates reported in Table \ref{tab:Flickr_overall}.}\label{tab:Flickr_perclass}
  \scriptsize
  \setlength{\tabcolsep}{3.5pt}
  \begin{tabular}{rcccccccc|c}
    \bottomrule
               & people & clothing & bodypart & animal & vehicle & instrument & scene & other & mean \\\hline
    MCG\_base  & 36.1 & 30.1 & 9.9  & 50.8 & 37.8 & 26.5 & 31.5 & 19.1 & 30.3  \\\hline
    Grad (MCG)       & 65.0 & 32.4 & 14.0 & \textbf{70.1} & 63.0 & 40.7 & 58.8 & 32.5 & 47.1  \\
    Deconv (MCG)    & 65.4 & 31.6 & 18.7 & 67.0 & 64.0 & 46.9 & 53.6 & 28.9 & 47.0  \\
    LRN (MCG)       & 64.6 & 37.7 & 16.4 & 62.9 & 63.5 & 45.7 & 59.4 & 37.9 & 48.5  \\
    CAM (MCG)       & 60.5 & 28.4 & 9.6  & 57.0 & 57.5 & 37.0 & \textbf{64.4} & 32.7 & 43.4  \\\hline
    MWP (MCG)      & \textbf{68.6}    & 37.7 & 16.1 & 68.7 & 66.3 & 53.7 & 54.5 & 36.8 & 50.3  \\
    c-MWP (MCG)      & 63.5    & \textbf{47.6} & \textbf{24.5} & 69.9 & \textbf{72.0} & \textbf{54.3} & 61.0 & \textbf{40.2} & \textbf{54.1}  \\\toprule\bottomrule
    CCA$^*$ \cite{plummer2015flickr30k} (EB)        & \textbf{63.6} & \textbf{43.7} & \textbf{22.9} & 57.0 & 69.0 & 50.6 & 45.0 & \textbf{36.2} & 48.5\\
    c-MWP (EB)  & 62.8 & 35.0 & 17.6 & \textbf{65.1} & \textbf{73.5} & \textbf{58.6} & \textbf{53.2} & \textbf{36.2} & \textbf{50.3}\\
    \toprule
  \end{tabular}
\end{table}

To generate scored bboxes for an NP, we first compute the word attention map for each word in the NP using our tag classifier. Images are resized to 300 pixels in the shortest dimension to better localize small objects. Then we simply average the word attention maps to get an NP attention map. Advanced language models can be used for better fusing the word attention maps, but we adopt the simplest fusion scheme to demonstrate the effectiveness of our top-down attention model.
We skip a small proportion of words that are not covered by our 18K dictionary. MCG \cite{APBMM2014} is used to generate 500 segment proposals, which are re-scored based on the phrase attention map. The re-scored segments are then converted to bboxes, and redundant bboxes are removed via Non-maximum Suppression using the IOU threshold of 0.7.

The segment scoring function is defined as $f(R) = S_R/A_R^\gamma$ where $S_R$ is the sum of the values inside the segment proposal $R$ on the given attention map and $A_R$ is the segment's area. The parameter $\gamma$ is to control the penalty of the segment's area, which is optimized for each method in the range $[0:0.25:1]$.

\textbf{Results.} The recall rates and mAP scores are reported in Table \ref{tab:Flickr_overall}. For our method and the baselines, we additionally report the performance on a subset of small instances whose bbox area is below $0.25$ of the image size, as we find small regions are much more difficult to localize. Our c-MWP consistently outperforms all the attention map baselines across different metrics. In particular, the group-level mAP of our method is better than the second best by a large margin.

We also compare with a recent fully supervised method \cite{plummer2015flickr30k}, which is trained directly on the Flickr30k Entities dataset using CNN features.
For fair comparison, we use the same bbox proposals used in \cite{plummer2015flickr30k}, which are generated by EdgeBoxes (EB) \cite{zitnick2014edge}. These proposals are pre-computed and provided by \cite{plummer2015flickr30k}. Our performance using EB is lower than using MCG, mainly due to the lower accuracy of the EB's bbox proposals. Compared with the segmentation proposals, the bbox proposals can also affect our ranking function for small and thin objects. However, our method still attains competitive performance against \cite{plummer2015flickr30k}. Note that our method is weakly supervised and does not use any training data from the Flickr30k Entities dataset.

\begin{figure}[t]
  \centering
  \includegraphics[width=1\linewidth]{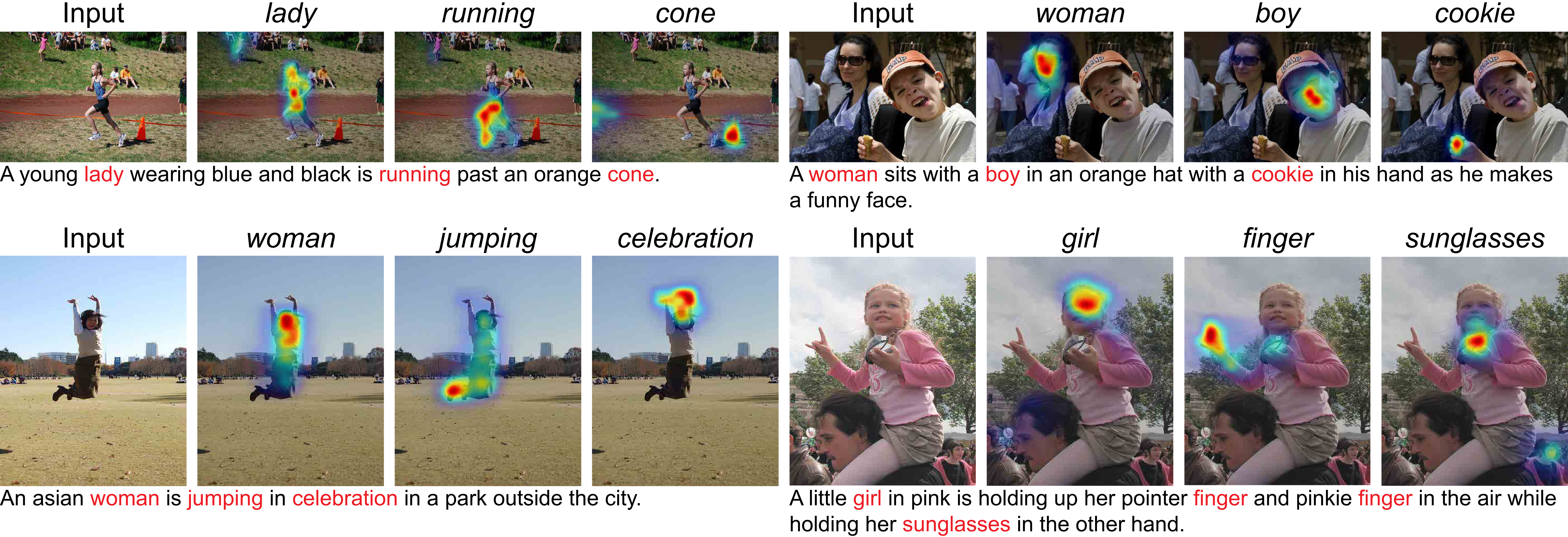}\\
  \caption[Example word attention maps obtained by c-MWP]{Word attention maps obtained by c-MWP using our image tag classifier. For each test image, one of its caption annotations from Flickr30k Entities is displayed below. We show the attention maps for the words in red in each caption. By leveraging a large-scale weakly labeled dataset, our method can localize a large number of visual concepts, \emph{e.g.} objects (cone, sunglasses and cookie), fine-grain categories of people (woman and boy), body parts (finger) and actions (jumping, running and celebration).}\label{fig:Flickr_demo}
\end{figure}
We further report the per-group Recall@5 score in Table \ref{tab:Flickr_perclass}. Our method achieves promising results in many group types, \emph{e.g.}\ \texttt{vehicle} and \texttt{instrument}. Note that the fully supervised CCA (EB) \cite{plummer2015flickr30k} gives significantly worse performance than c-MWP (EB) in \texttt{animal}, \texttt{vehicle} and \texttt{instrument}, which are the three rarest types in the Flickr30k Entities dataset. This again shows the limitation of fully-supervised approaches due to the lack of fully-annotated data.

Some example word attention maps are shown in Fig.~\ref{fig:Flickr_demo} to demonstrate the localization ability of our method. As we can see, our method can localize not only noun phrases but also actions verbs in the text.

\section{Conclusion}

We propose a probabilistic Winner-Take-All formulation to model the top-down neural attention for CNN classifiers. Based on our formulation, a novel propagation method, Excitation Backprop, is presented to compute the Marginal Winning Probability of each neuron. Using Excitation Backprop, highly discriminative attention maps can be efficiently computed by propagating a pair of contrastive top-down signals via a single backward pass in the network. We demonstrate the accuracy and the generalizability of our method in a large-scale Pointing Game. We further show the usefulness of our method in localizing dominant objects. Moreover, without using any localization supervision or language model, our neural attention based method attains competitive localization performance \emph{vs.}\ a state-of-the-art fully supervised method on the challenging Flickr30k Entities dataset.

\textbf{Acknowledgments.} This research was supported in part by Adobe Research, US NSF grants 0910908 and 1029430, and gifts from NVIDIA.

\bibliographystyle{splncs}
\bibliography{egbib}

\newpage
\begin{appendix}
\section*{Appendix}
\section{Speed Performance} \label{sec:EB2}
\begin{figure}[h]
  \centering
  \includegraphics[width=0.55\linewidth]{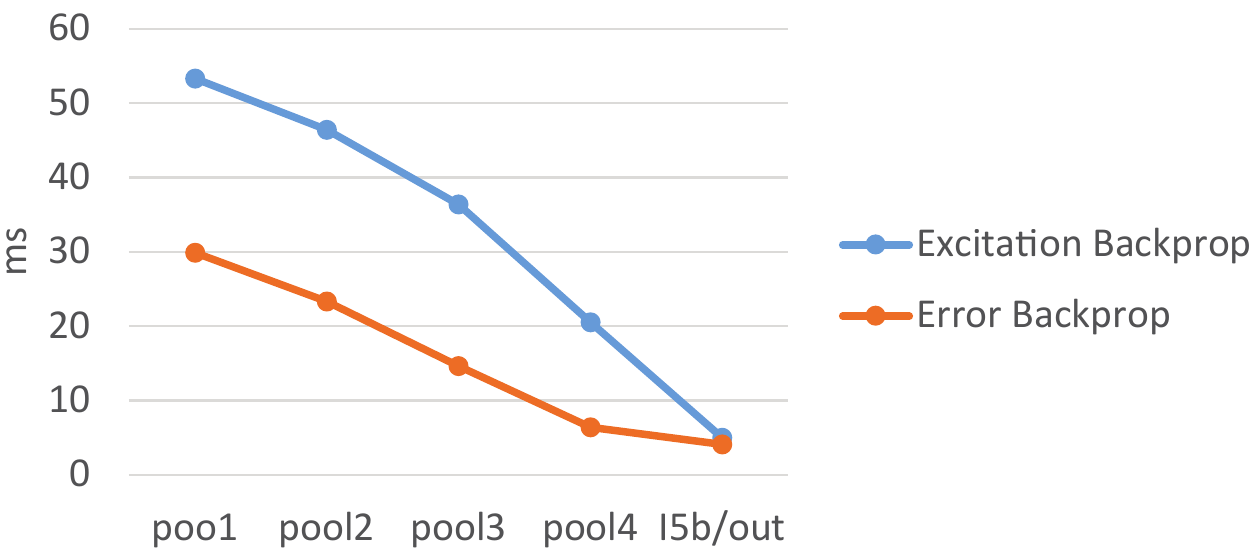}\\
  \caption{Speed performance of our implementation of Excitation Backprop compared with error backpropagation in GPU mode. The speed is measured on a NVIDIA K40c GPU for a single 224X335 image (without using batch mode). The x-axis represents the layer at which the tested method terminates.}\label{fig:speed}
\end{figure}

The most time-consuming operations in Excitation Backprop are the second and fourth steps in Alg.~\ref{alg:EB}, which correspond to the \texttt{forward} and \texttt{backward} operations of the layer in Caffe. Therefore, the computational complexity of Excitation Backprop is about twice the complexity of error backpropagation, but in practice we only perform the Excitation Backprop to some intermediate layer. The speed performance of our implementation of Excitation Backprop is reported in Fig.~\ref{fig:speed} for \texttt{GoogleNet}.
\section{Details about the Stock6M Dataset}\label{sec:TR1}

We provide more details about the Stock6M dataset used for training the image tag classifier.

\textbf{Data collection and cleaning.} We crawl an initial set of about 17M thumbnail images and their tags from a stock image website. This website provides professional photos and illustrations for commercial usage. Each image on the website has a list of tags used for text-based image search. Then we use the most frequent 18157 tags for our dictionary using a frequency threshold of 1000. Most of these tags are unigrams. We remove images with fewer than five tags. We empirically find that some images' tags are in alphabetical order, and the quality of these tags is usually poor. Thus, we remove these images, too. We further perform a duplicate detection based on tag information and user ids, since many images uploaded by the same user can be very similar. For each user id, we check the tag list of each of its images. We remove an image if its first five tags are very similar to the first five tags of  a previously seen image of the same user id. The two sets of tags are considered to be similar if they have more than three overlaps. After all these steps, we end up with a dataset of about 6M images.

\begin{figure}[t]
  \centering
  \includegraphics[width=1\linewidth]{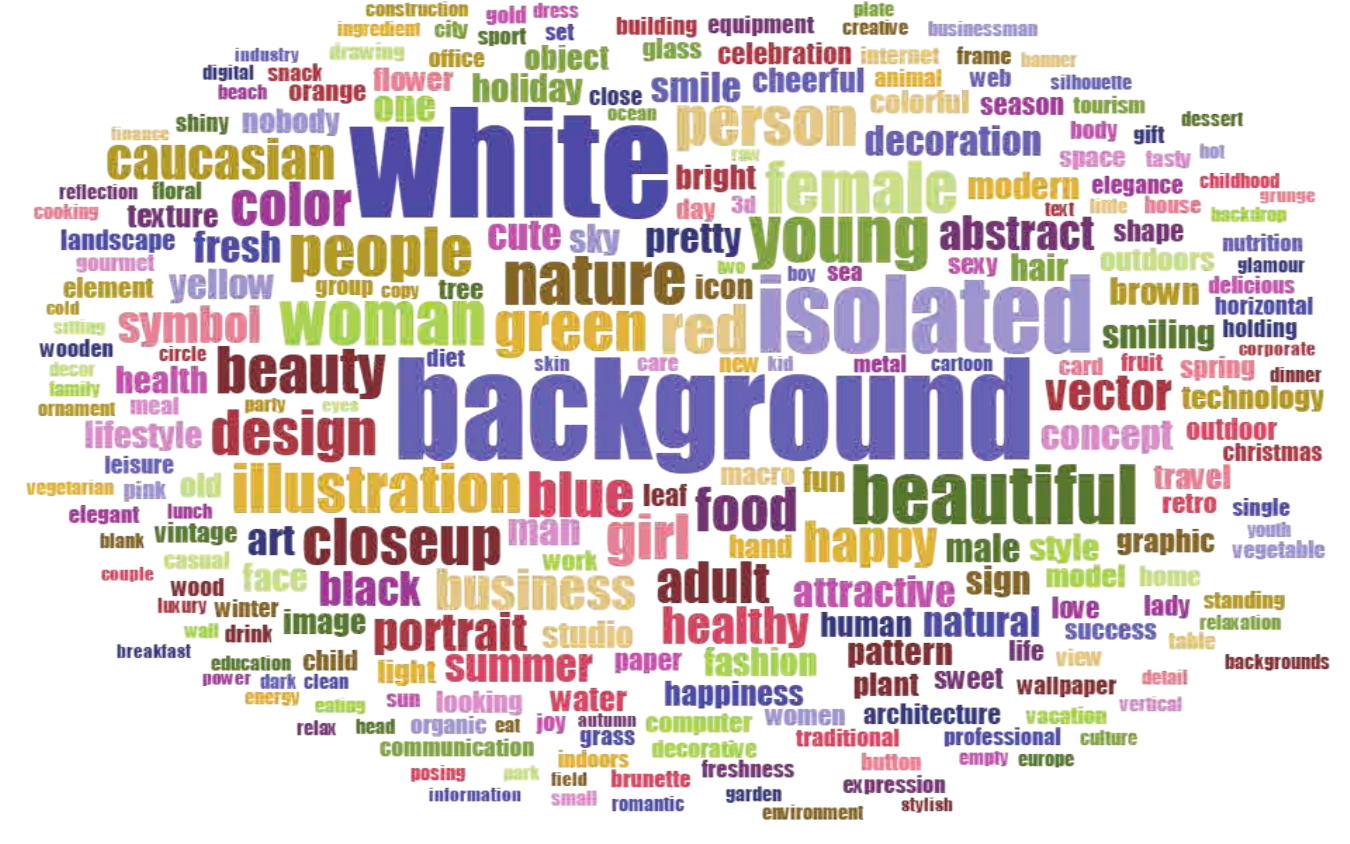}\\
  \caption{Stock6M: visualization of the most frequent tags in Stock6M.}\label{fig:wc}
\end{figure}

\textbf{Frequent Tags and Example Images.} We visualize the most frequent tags in a word cloud (Fig.~\ref{fig:wc}). We can see that many frequent tags are related to humans, for example \texttt{woman}, \texttt{man}, \texttt{beautiful}, \texttt{happy}, \emph{etc}. There are also a lot of non-visual tags like \texttt{healthy}, \texttt{business}, \texttt{holiday} and \texttt{lifestyle}. Some example images and the corresponding user tags are shown in Fig~\ref{fig:example}.

\begin{figure}[t!]
  \centering
  \includegraphics[width=1\linewidth]{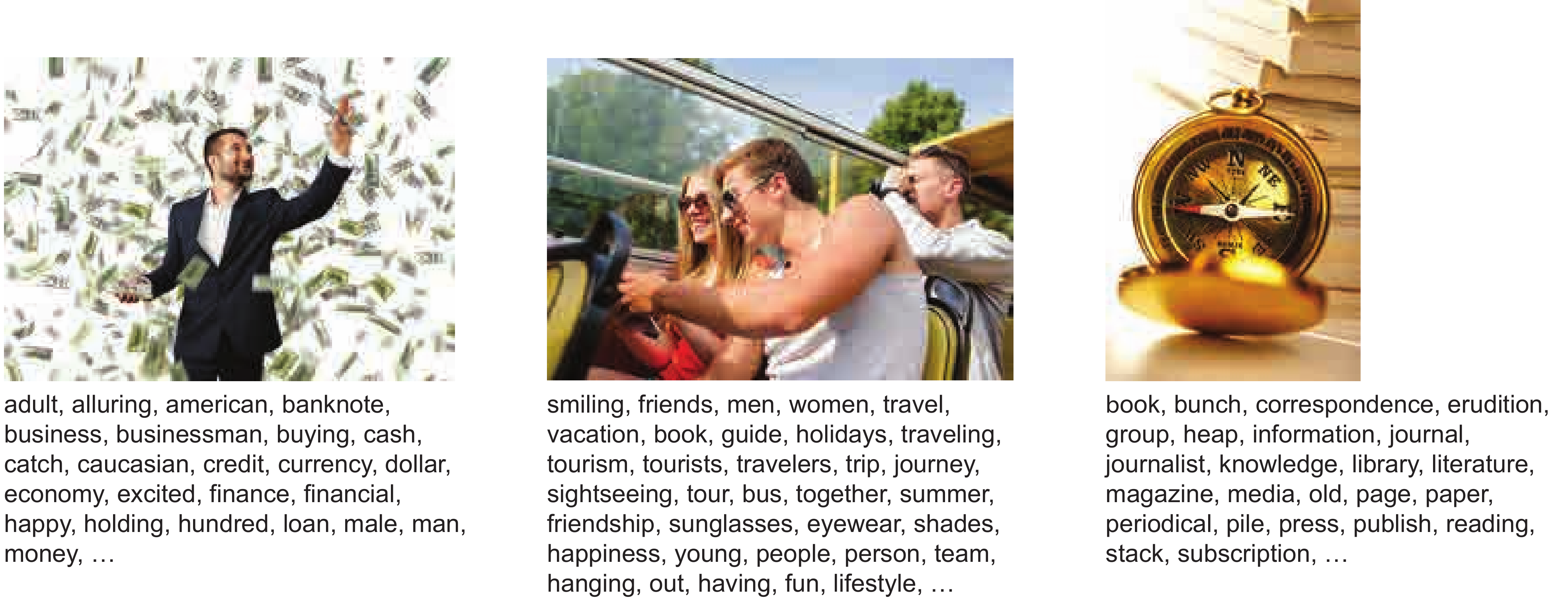}\\
  \caption{Stock6M: example images and tags from the stock image website. Note that the first and third images' tags are in alphabetical order. In contrast, the second image's tags are not in alphabetical order and their ordering roughly reflects the relevance to the image. Therefore, we remove images whose tags are in alphabetical order.}\label{fig:example}
\end{figure}

\end{appendix}
\end{document}